\RequirePackage{fix-cm}
\documentclass[twocolumn]{svjour3}          
\smartqed  
\usepackage{multirow}
\usepackage{makecell}
\usepackage{times}
\usepackage{soul}
\usepackage{url}
\usepackage[utf8]{inputenc}
\usepackage{graphicx}
\usepackage{amsmath}
\usepackage{booktabs}
\usepackage{algorithm, algpseudocode}
\usepackage{epsfig}
\usepackage{graphicx}
\usepackage{amssymb}
\usepackage{array}
\date{} 
\usepackage[T1]{fontenc}    
\usepackage{url}            
\usepackage{booktabs}       
\usepackage{amsfonts}       
\usepackage{nicefrac}       
\usepackage{microtype}      
\usepackage{comment}
\usepackage{colortbl}
\usepackage{xfrac}
\usepackage[outercaption]{sidecap}

\algrenewcommand\algorithmicrequire{\textbf{Input:}}
\algrenewcommand\algorithmicensure{\textbf{Output:}}

\newcolumntype{M}[1]{>{\centering\arraybackslash}m{#1}}

\def\eg{\emph{e.g.}} 
\def\ie{\emph{i.e.}}

\newlength\savewidth\newcommand\shline{\noalign{\global\savewidth\arrayrulewidth
  \global\arrayrulewidth 1pt}\hline\noalign{\global\arrayrulewidth\savewidth}}

\makeatletter
\renewcommand*\env@matrix[1][\arraystretch]{%
  \edef\arraystretch{#1}%
  \hskip -\arraycolsep
  \let\@ifnextchar\new@ifnextchar
  \array{*\c@MaxMatrixCols c}}
\makeatother

\usepackage{cite}
\usepackage{natbib}
\usepackage[dvipsnames,svgnames,x11names]{xcolor}
\definecolor{citecolor}{RGB}{119,185,0} 
\usepackage[pagebackref=false,breaklinks=true,letterpaper=true,colorlinks,citecolor=citecolor,bookmarks=false]{hyperref}

\begin{document}

\title{ Learning Cross-view Geo-localization Embeddings via Dynamic Weighted Decorrelation Regularization}

\author{
  Tingyu Wang, Zhedong Zheng, Zunjie Zhu, Yuhan Gao, Yi Yang and Chenggang Yan \\ 
  }
\institute{
T. Wang, C. Yan are with the Intelligent Information Processing Lab, Hangzhou Dianzi University, China 310018. E-mail: wongtyu@hdu.edu.cn, cgyan@hdu.edu.cn. C. Yan is the Corresponding Author. \\
Z. Zheng is with the Sea-NExT Joint Lab, School of Computing, National University of Singapore, Singapore 118404. E-mail: zdzheng@nus.edu.sg. \\
Z. Zhu is with the Intelligent Information Processing Lab, Hangzhou Dianzi University, China 310018, also with the Lishui Institute of Hangzhou Dianzi University, China 323000. E-mail: zunjiezhu@hdu.edu.cn. \\
Y. Gao is with the Lishui Institute of Hangzhou Dianzi University, China 323000, also with the Intelligent Information Processing Lab, Hangzhou Dianzi University, China 310018. E-mail: yuhangao@hdu.edu.cn. \\
Y.Yang is with the School of Computer Science, Zhejiang University, China, 310027. E-mail:~yangyics@zju.edu.cn.
}

\maketitle
\begin{abstract}
Cross-view geo-localization aims to spot images of the same location shot from two platforms, \eg, the drone platform and the satellite platform. 
Existing methods usually focus on optimizing the distance between one embedding with others in the feature space, while neglecting the redundancy of the embedding itself. 
In this paper, we argue that the low redundancy is also of importance, which motivates the model to mine more diverse patterns.  
To verify this point, we introduce a simple yet effective regularization, \ie, Dynamic Weighted Decorrelation Regularization (DWDR), to explicitly encourage networks to learn independent embedding channels. 
As the name implies, DWDR regresses the embedding correlation coefficient matrix to a sparse matrix, \ie, the identity matrix, with dynamic weights. 
The dynamic weights are applied to focus on still correlated channels during training. 
Besides, we propose a cross-view symmetric sampling strategy, which keeps the example balance between different platforms. Albeit simple, the proposed method has achieved competitive results on three large-scale benchmarks, \ie, University-1652, CVUSA and CVACT. Moreover, under the harsh circumstance, \eg, the extremely short feature of 64 dimensions, the proposed method surpasses the baseline model by a clear margin.
\end{abstract}

\begin{figure}[tb]
  \centering
  \includegraphics[width=0.95\linewidth]{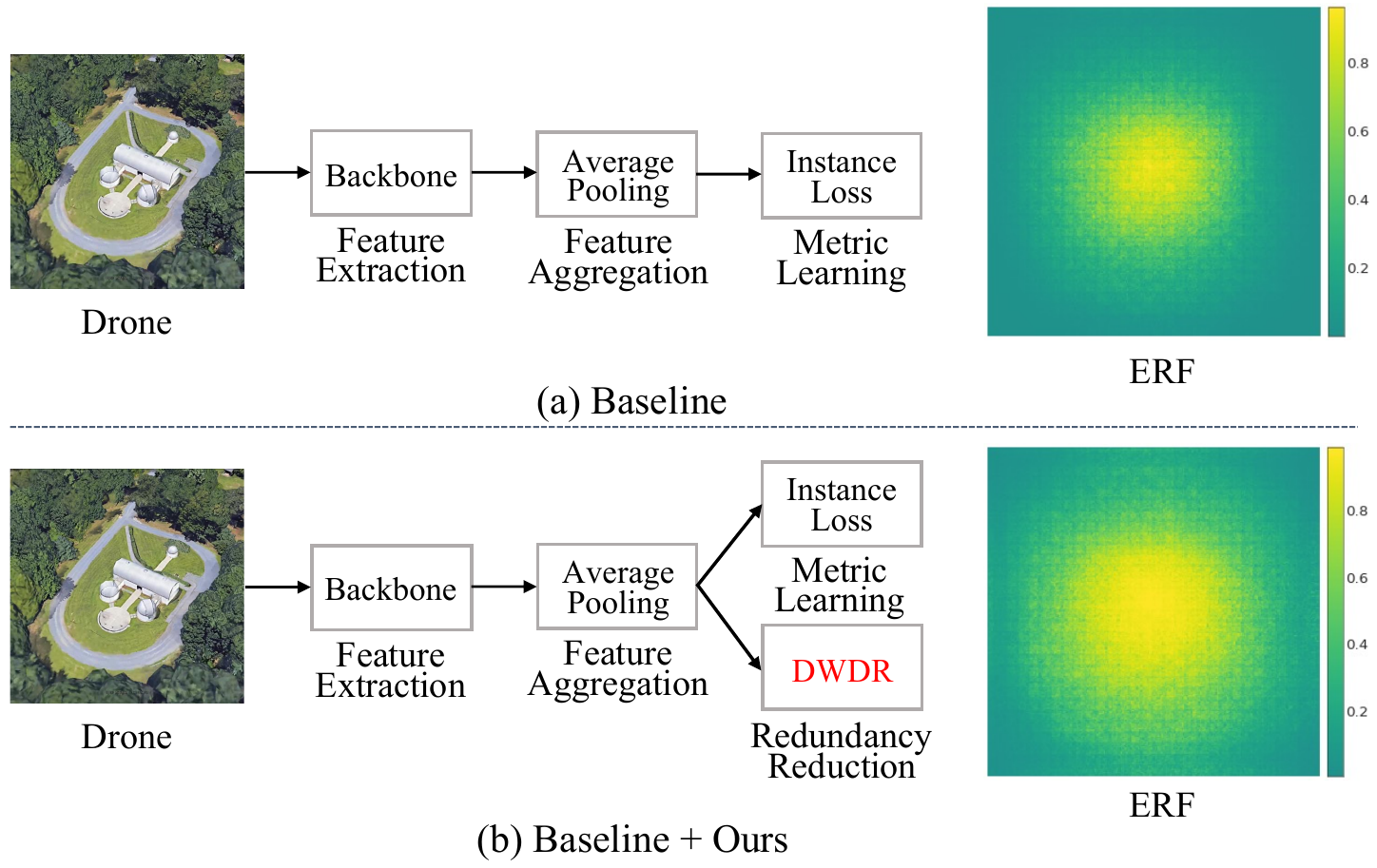}
  \caption{(a) A strong metric learning based baseline~\citep{zheng_university-1652_nodate}. (b) The baseline combined with our proposed dynamic weighted decorrelation regularization (DWDR). ERF refers to the effective receptive field (the yellow area). A large ERF reflects that the learned model is able to extract more discriminative visual features from a wider range without over-fitting to the local area~\citep{luo2016understanding,ding2022scaling}. 
  }
  \label{fig:1}
\end{figure}

\section{Introduction}
Cross-view geo-localization is an image retrieval task and has been broadly applied to event detection, drone navigation, and accuracy delivery~\citep{zheng_university-1652_nodate, wang2021each,humenberger2022investigating}.
Given a probe from the query platform (\eg, the drone), the system aims to spot a candidate image in the gallery platform (\eg, the satellite) containing the same geographic target with the probe. Since satellite images possess GPS metadata as annotations, we can easily acquire the location information of the interesting probe. In addition, when the GPS signal of the positioning device encounters interference, the image-based cross-view geo-localization can also be employed as an auxiliary tool to refine the geo-localization and provide a more robust result. 
\par
Cross-view geo-localization remains challenging since images from different platforms inherently contain viewpoint variations. The extreme viewpoint change leads to differences in the visual appearance of a geographic target, which confuses the system to locate a position accurately. A crucial key to the geo-localization challenge is to learn a discriminative visual embedding~\citep{teng2021viewpoint,zheng2020vehiclenet,huang2022learning}. Recently, deep learning technologies have received much research attention in the cross-view geo-localization problem since the great potential in feature extraction. A popular scheme for learning cross-view geo-localization models is first utilizing the pre-trained convolutional neural network to extract feature maps of images. Following, various metric learning functions are proposed to pull the image pairs with the same geo-tag closer while pushing those features from non-matchable pairs far apart~\citep{zheng_university-1652_nodate,liu_lending_2019,hu_cvm-net_2018}. Based on this basic scheme, the attention mechanism~\citep{lin2022joint, shi_spatial-aware_nodate} and aligning the spatial layout of features~\citep{wang2021each, liu_lending_2019, shi_spatial-aware_nodate, shi_optimal_nodate, hu2020image} are also widely considered in the network design. Most existing methods focus more on the similarity between cross-view embeddings while ignoring the redundant channels of the embedding itself.
\par
In human perception study, the neuroscience H. Barlow claims that concise, non-redundant descriptions are of higher value to the perception system and will help to clarify inputs of the external world~\citep{barlow1961possible}.
Based on this bio-perceptual hypothesis, we argue that stripping the redundancy of visual embeddings contributes to the discrimination of different targets. 
In this paper, we propose a dynamic weighted decorrelation regularization (DWDR). 
The measuring objective of DWDR is the Pearson cross-correlation coefficient matrix, which is computing from features of positive pairs composed of cross-view images. 
Specifically, DWDR employs Square Loss to regress the diagonal elements of the objective matrix to 1, and the off-diagonal elements are approximated to 0.
However, the objective matrix is typically large, \eg, $2048 \times 2048$.
The optimizer is easily overwhelmed by a mass of elements already close to the optimization goal, thus ignoring other elements that need to be regressed. 
To address this optimization problem, we assign a dynamic weight to each element loss according to the maximum regression distance of the target element.  
The dynamic weight can adaptively highlight the importance of poorly-regressed elements and suppress the side effect of well-regressed elements in optimization.
We observe that DWDR encourages the learned model to focus on a larger effective receptive field, which prevents the network from overfitting to a local pattern~\citep{luo2016understanding} (see Figure~\ref{fig:1}). Besides, as mentioned above, the computation of the Pearson cross-correlation coefficient matrix requires positive sample pairs. As a by-product of selecting positive pairs, we also provide a cross-view symmetric sampling strategy. In a training batch, our symmetric sampling strategy aligns the number of the same geo-tag images between different platforms. Therefore, the proposed strategy mitigates the sample imbalance, especially in drone-to-satellite geo-localization, which contains limited satellite data~\citep{ding2021practical, dai2021transformer}.
To summarize, our contributions are as follows.
\begin{itemize}
\setlength\itemsep{0em}
\item We propose a dynamic weighted decorrelation regularization (DWDR), which motivates networks to learn discriminative embeddings by stripping the redundancy of features. During training, DWDR assigns a dynamic weight to the loss of each element in the objective matrix, yielding efficient optimization of networks. As a by-product of DWDR, we further introduce a cross-view symmetric sampling strategy, which maintains the example balance in a training batch.
\item Albeit simple, we demonstrate the effectiveness of the proposed method on three cross-view geo-localization datasets.
Extensive experiments show that multiple existing works~\citep{zheng_university-1652_nodate,wang2021each, liu2021swin} fused with our method are able to further boost the performance. In addition, we observe that our method still obtains superior results even for a short visual embedding with 64 dimensions.
\end{itemize}
\par
The rest of this paper is organized as follows. In Section~\ref{related_work}, we discuss related works. The details of our method are illustrated in Section~\ref{proposed method}. Experimental results are provided in Section~\ref{experiment}. Finally, Section~\ref{conclusion} presents a summary.


\section{Related Work} \label{related_work}
In this section, we briefly review related previous works, including image-based cross-view geo-localization and low-redundancy representation learning.
\subsection{Image-based Cross-view Geo-localization}
Imaged-based cross-view geo-localization has been tackled as an image retrieval task. Early works~\citep{bansal2011geo, lin2013cross, SemanticCM,saurer2016image} employ hand-crafted operators to extract discriminative features for cross-view image matching. With the development of deep learning, the convolutional neural network (CNN) has received much research attention in the extraction of image representation. The pioneering CNN-based approach~\citep{workman_location_2015} directly deploys pre-trained AlexNet~\citep{krizhevsky2012imagenet} to extract features for the cross-view geo-localization. Further, \cite{workman_wide-area_2015} introduce the information of image pairs as the constraint to fine-tune the pre-trained network and acquire a better performance. Following this line of considering object constraints,~\cite{lin_learning_2015} borrow knowledge from face verification and harness the contrastive loss~\citep{hadsell2006dimensionality} to guide the optimization of a modified Siamese Network~\citep{chopra2005learning}. \cite{vo_localizing_2017} discuss the limitation of the Siamese architecture in large-scale cross-view matching and provide a soft-margin triplet loss to improve the geo-localization accuracy. Similarly, \cite{hu_cvm-net_2018} propose a weighted soft-margin ranking loss, which not only improves the matching accuracy but also speeds up the training convergence.~\cite{Siam-FCANet} mine hard examples in the training batch to strengthen the penalization of the soft-margin triplet loss.~\cite{zheng_university-1652_nodate} suggest that images with the same identification can be classified into one cluster and apply the instance loss~\citep{zheng2020dual,zheng2017unlabeled} as the proxy target to learn discriminative embeddings. Another line of works concentrates on addressing the spatial misalignment problem of cross-view retrieval. CVM-Net~\citep{hu_cvm-net_2018} employs a shared NetVLAD to aggregate the local feature to minor the visual gap between different viewpoints. \cite{liu_lending_2019} explicitly encode the orientation information into the image descriptors and boost the discriminative power of the learned features. \cite{shi_optimal_nodate} first attempt to utilize the optimal transport (OT) theory to close the spatial layout information in the high-level feature. Then \cite{shi_spatial-aware_nodate} directly resort to the polar transform to align the pixel-level semantic information of cross-view images. DSM~\citep{Shi_2020_CVPR} designs a dynamic similarity matching module to solve the cross-view matching in a limited Field of View (FoV). LPN~\citep{wang2021each} stresses the importance of contextual information and proposes a square-ring partition strategy to improve the performance of cross-view geo-localization. Without any extra annotations, RK-Net~\citep{lin2022joint} automatically detects salient keypoints to improve the model capability against the appearance changes.

\subsection{Low-Redundancy Representation Learning}
In the early study of human perception, the neuroscientist H. Barlow~\citep{barlow1961possible} suggests that the perception system tends to encode the raw sensory input as the low-redundancy representation in which each component possesses statistical independence. This learning principle guides a number of algorithms in machine learning. \cite{bengio2009slow} support that the decorrelation criterion is useful in the context of data and derive a fast online pretraining algorithm to learn decorrelated features for neural networks. \cite{cogswell2015reducing} design Decov Loss that motivates the network to learn non-redundant representations and demonstrate that decorrelating representations helps to reduce overfitting of the trained deep networks. \cite{sun2017svdnet} utilize Singular Vector Decomposition (SVD) and reduce the correlation between output units by integrating the orthogonality constraint in CNN training. Thus the final descriptor contains lower redundant information about the sample. In self-supervised learning (SSL), Barlow Twins~\citep{zbontar2021barlow} proposes a simple yet effective object function to acquire representations with low redundancy and avoid model collapse. The optimization goal of Barlow Twins is to transform a cross-correlation matrix into an identity matrix. SSL does not require the input data with human annotation, and the cross-correlation matrix is computed from two distorted representations of a sample.

\begin{figure*}[htb]
  \centering
  \includegraphics[width=0.95\linewidth]{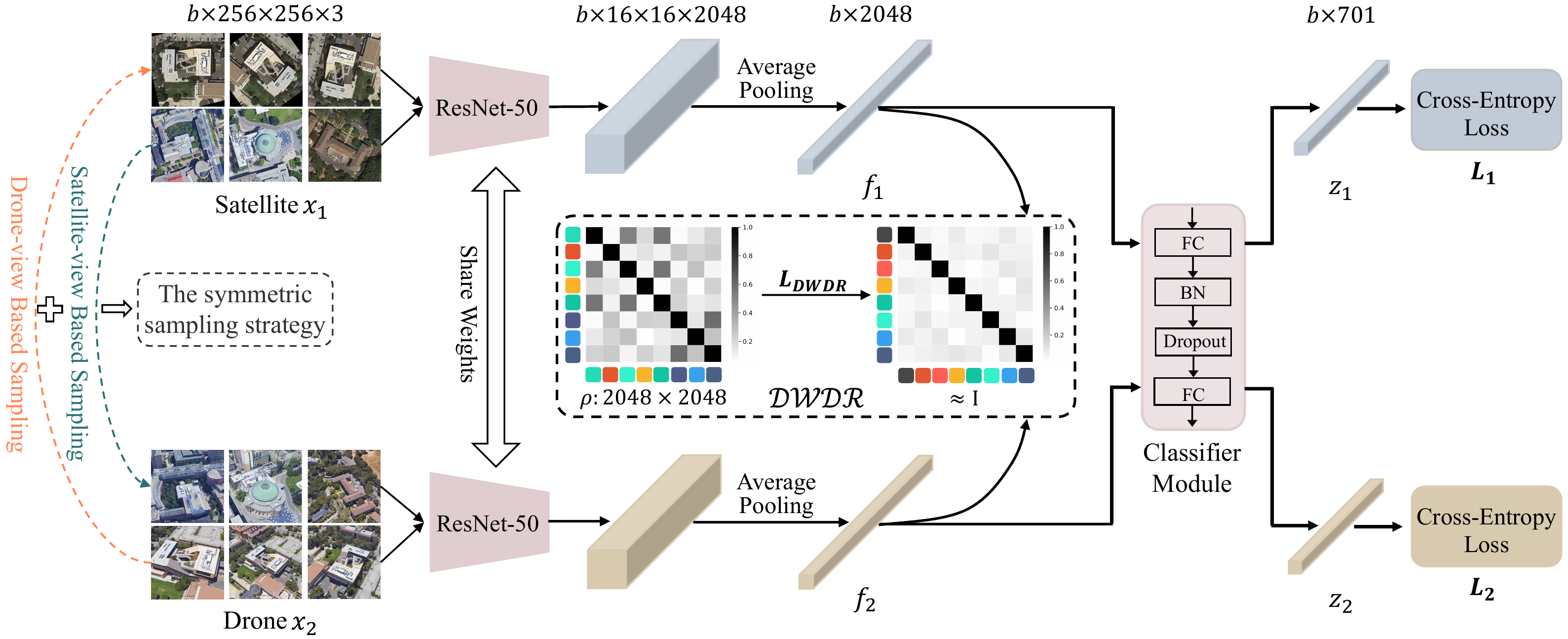}
  \caption{ A schematic overview of our method. We first apply the symmetric sampling strategy to generate one batch of drone-and-satellite positive pairs. The symmetric sampling strategy is composed of the drone-view based sampling and the satellite-view based sampling. Then the satellite-platform and the drone-platform images of the batch are fed into a two-branch network. The two branch network shares weights of two backbones since images from the drone platform and the satellite platform have similar patterns. Next, the average pooling is deployed to aggregate the output feature maps of each branch into the column vectors. Finally, the column vectors of two branches are inputted into a classifier module to acquire predicted logit scores, respectively, and the cross-entropy function is utilized to compute the instance loss. The proposed DWDR aims to transform the Pearson cross-correlation coefficient matrix $\rho$ into an identity matrix $I$ as much as possible. The Pearson cross-correlation coefficient matrix $\rho$ is calculated by column vectors from two branches. Note that here we show the framework employing ResNet-50 as the backbone and images of University-1652 as inputs.
  }
  \label{fig:2}
\end{figure*}

\section{Proposed Method} \label{proposed method}
In Section~\ref{pre}, we first give a revisit of preliminaries followed by the description of our baseline network structure for geo-localization in Section~\ref{sec:network}. Next, we introduce the dynamic weighted decorrelation regularization (DWDR). The dynamic weight mechanism relieves the plateau problem in Barlow Twins~\citep{zbontar2021barlow}. We also provide a mechanism discussion (see Section~\ref{sec:dwdr}). 

\subsection{Preliminaries}~\label{pre}
\label{revisit}
\textbf{Cross-correlation matrix} measures the correlation between two matrices. In particular, for two random matrices $\mathbf{X}=(X_{1}, X_{2},\cdots, X_{M})^{T}$, $\mathbf{Y} = (Y_{1},Y_{2} \cdots, Y_{N})^{T}$, where $X_m,Y_n \in \mathbb{R}^{a}$ with $a$ dimensions, the cross-correlation matrix of $\mathbf{X}$ and $\mathbf{Y}$ can be defined as:
\begin{align}\label{func1}
    \phi &\triangleq \mathbf{\mathbb{E}\left[XY^{T}\right]}.
\end{align}
A component-wise description is:
\begin{align}\label{func2}
\setlength{\arraycolsep}{0.8pt}
\begin{array}{lc}
    \phi &= \begin{bmatrix}[1.2]
                        \mathbb{E}\left[X_{1}Y_{1}^{T}\right] & \mathbb{E}\left[X_{1}Y_{2}^{T}\right] & \cdots & \mathbb{E}\left[X_{1}Y_{N}^{T}\right] \\
                        \mathbb{E}\left[X_{2}Y_{1}^{T}\right] & \mathbb{E}\left[X_{2}Y_{2}^{T}\right] & \cdots & \mathbb{E}\left[X_{2}Y_{N}^{T}\right] \\
                        \vdots & \vdots & \ddots & \vdots \\
                        \mathbb{E}\left[X_{M}Y_{1}^{T}\right] & \mathbb{E}\left[X_{M}Y_{2}^{T}\right] & \cdots & \mathbb{E}\left[X_{M}Y_{N}^{T}\right] \\
                        \end{bmatrix},
\end{array}
\end{align}
where $\mathbb{E\left[\cdot\right]}$ refers to the expectation.
\par
\textbf{Pearson cross-correlation coefficient matrix} is similar to the cross-correlation matrix, and it shows a normalized measurement between two matrices. Differently, the value of each element in the Pearson cross-correlation coefficient matrix is between $-1$ and $1$. The score $1$ and $-1$ denote that two vectors are perfectly correlated and anti-correlated. $0$ means that two vectors are completely unrelated. Based on the definition of the cross-correlation matrix, the Pearson cross-correlation coefficient matrix can be formulated as:
\begin{align}\label{func3}
    \rho &= \frac{\mathbb{E}\left[\left(\mathbf{X - \mu_{X}}\right)\left(\mathbf{Y - \mu_{Y}}\right)^{T}\right]}{\sigma_{\mathbf{X}}\sigma_{\mathbf{Y}}},
\end{align}
where $\mu_{\mathbf{X}}$ and $\sigma_{\mathbf{X}}$ are the mean and the standard deviation of $\mathbf{X}$, respectively. $\mu_{\mathbf{Y}}$ and $\sigma_{\mathbf{Y}}$ are the mean and the standard deviation of $\mathbf{Y}$, separately. $\mathbb{E\left[\cdot\right]}$ is the expectation.
\par
\textbf{Barlow Twins}~\citep{zbontar2021barlow} is a widely-used method for self-supervised learning (SSL). To address the issue of representation collapse, the main contribution of Barlow Twins is introducing a new regularization objective, which can be defined as:
\begin{equation}
L_{BT} \triangleq \sum_{i=1}^d (1-\phi_{ii})^2 + \lambda \sum_{i=1}^d\sum_{j=1,j \neq i}^d {\phi_{ij}}^2,
\label{eq:lossBarlow}
\end{equation}
where $\lambda$ is a positive hyper-parameter, and $\phi$ is the cross-correlation matrix between two mini-batch of features. Given two batches of features with $d$ dimensions ($d$ is generally set as 2048), we multiply the feature matrix along with the batch dimension. Thus, the dimension of $\phi$ is $d \times d$. The regularization function $L_{BT}$ encourages every feature channel to be independent of others. Specifically, it impels the diagonal elements $\phi_{ii}$ from the same channel to $1$, while pushing off-diagonal elements $\phi_{ij}$ between different channels to $0$. 
However, in practice, Barlow Twins meets the optimization problem, especially when facing a typical large matrix (\eg, $2048 \times 2048$). The model arrives at the plateau after the majority of channels are converged, and it neglects other still-correlated ``hard'' channels, compromising the training process.   

\subsection{Network Structure}\label{sec:network}
We adopt ResNet-50~\citep{he2016deep} as the backbone and add a new classifier. The classifier consists of a fully-connected layer (FC), a batch normalization layer (BN), a dropout layer (Dropout), and another fully-connected layer (FC). 
Notably, the backbone can also be other networks such as VGG16~\citep{vgg} and Swin Transformer~\citep{liu2021swin}. 
The two-branch baseline consists of three forward phases, \ie, feature extraction, feature aggregation, and feature classification. 
Specifically, we denote the input images from two platforms as $x_{k}$, where $k \in \{1, 2\}$. $1$ denotes the satellite platform, and $2$ refers to the drone or ground platform. 
We first employ two backbones with shared weights 
to extract feature maps. 
Then the global average pooling is deployed to aggregate the information of feature maps into the column vectors $f_{k}$. Finally, we harness a classifier to map vectors $f_{k}$ of different platforms into one shared classification space and acquire the predicted logit vectors $z_{k}$. 
Meanwhile, the cross entropy function is employed to calculate the instance loss $L_{id}$ ~\citep{zheng_university-1652_nodate}. The instance loss is a classification loss with a shared classifier $\mathcal{F}_{classifier}$.
The above process can be formulated as:
\begin{align}
 f_{k} &= \mathcal{A}vgpool(\mathcal{F}^k_{backbone}(x_{k})),
\label{func4}
\end{align}
\begin{align}
 z_{k} &= \mathcal{F}_{classifier}(f_{k}),
\label{func6}
\end{align}
\begin{align}
 L_{id} &= \sum_{k=1}^{2}-log\frac{exp(z_{k}(y))}{\sum_{c=1}^{C}exp(z_{k}(c))}.
\label{func7}
\end{align}
The label $y \in [1, C]$, where $C$ indicates the category number of geographic targets in the training set.
$z_{k}(y)$ is the logit score of the ground-truth geo-tag $y$. When inference, we remove the final linear classification layer and extract the intermediate feature $f_k$ as the visual representation.
 
\subsection{Dynamic Weighted Decorrelation Regularization} \label{sec:dwdr}
In this work, we introduce a dynamic weighted decorrelation regularization (DWDR) to encourage the network to learn low-redundancy visual embeddings. 
As shown in Figure~\ref{fig:2}, DWDR is implemented based on a classic two-branch baseline~\citep{zheng_university-1652_nodate}. The two-branch baseline harnesses location classification as the pretext~\citep{zheng_university-1652_nodate,zheng2019joint} to conduct the cross-view geo-localization task. During training, we employ the symmetric sampling strategy to balance examples between different platforms in a training batch. It is worth noting that the symmetric sampling strategy is a by-product of DWDR.

The optimization objective of DWDR is the Pearson cross-correlation coefficient matrix $\rho$ between $f_{k}$ extracted from images of different platforms. Given two batches of extracted vectors $f_{1}$ and $f_{2}$ of size $b \times 2048$, according to Eq.~\ref{func3}, we can gain the objective matrix $\rho$ with the shape of $2048 \times 2048$, where $b$ denotes the batchsize. DWDR aims to spur the network by regressing the objective matrix $\rho$ into a sparse matrix, \ie, an identity matrix. We employ Square Loss to constrain the regression of each element. DWDR can be written as:
\begin{equation}
\begin{aligned}
L_{DWDR} &\triangleq \sum_{i=1}^d \omega_{1} \cdot (1-\mathcal{\rho}_{ii})^2 + \lambda \sum_{i=1}^d\sum_{j=1,j \neq i}^d \omega_2 \cdot \mathcal{\rho}_{ij}^2,
\label{eq:dwdr}
\end{aligned}
\end{equation}
where $\lambda$ is a hyper-parameter to balance the diagonal and off-diagonal element weight, $\rho_{ii}$ refers to the diagonal elements of the objective Pearson matrix $\rho$, and $\rho_{ij}$ denotes off-diagonal elements. 
$\rho_{ii}$ is regressed to $1$, which makes visual embeddings of the same geo-tag invariant for different platforms. 
$\rho_{ij}$ is regressed to $0$ to make the visual embedding channels independent from each other. 
$\omega_1$ and $\omega_2$ are two dynamic weights to prevent the optimization plateau, which depends on the regression score. In this way, the dynamic weight adjusts the influence of the loss and adaptively pays attention to the poorly-regressed elements during training. Considering that each element of the Pearson matrix is in $[-1,1]$, we set $\omega_{1} = \left(\frac{1-\rho_{ii}}{2}\right)^{\gamma_{1}}, \omega_{2} = \lvert\rho_{ij}\rvert^{\gamma_{2}}$, $\lvert \cdot \rvert$ denotes the absolute value.  
In this way, given non-negative focusing parameters $\gamma_{1}$ and $\gamma_{2}$, we ensure $\omega_{1} \in [0,1]$ and $\omega_{2} \in [0,1]$.
For elements close to the regression result (\textbf{well-regressed elements}),
the assigned dynamic weight is near $0$. 
Conversely, for elements far from the regression target (\textbf{poorly-regressed elements}), the assigned dynamic weight increases to $1$. 
\par
\textbf{Symmetric sampling strategy.} In order to compute the Pearson cross-correlation coefficient matrix, a necessary step is to construct a training batch by acquiring different-platform images with same geo-tags (`positive pairs').
Here we take the University-1652 dataset~\citep{zheng_university-1652_nodate} as an example. We can first randomly sample the satellite-view image as an anchor and then form the positive pair by finding the drone-view image of the same geo-tag. We call this sampling, which takes the satellite-view image as an anchor, the satellite-view based sampling. If we use the drone-view image as the anchor, we call the strategy the drone-view based sampling.
We note that the image number of different platforms is usually different due to the capturing difficulty. For example, we can easily acquire multiple drone images while only having one satellite image. If we apply the satellite-view based sampling during training, it will miss sampling some drone-view images. It is because every time we randomly sample one image from 54 drone-view images with putting back. 
On the other hand, if we apply the drone-view based sampling, it will contain duplicate geo-tags within a mini-batch, which repeatedly samples the same satellite image.  
Therefore, we propose a symmetric sampling strategy, which combines the satellite-view based sampling and drone-view based sampling. In particular, we sample two mini-batch by the two strategies respectively and combine them together as the new mini-batch to train the model. 
This combined strategy ensures the model can ``see'' all the training data while keeping the category sampling relatively balanced.  
\par


\textbf{Discussion.} 
The proposed method is similar to Barlow Twins considering disentangling the correlation matrix, but is different in the following aspects:
First, Barlow Twins~\citep{zbontar2021barlow} optimizes the cross-correlation matrix, while our method harnesses the Pearson cross-correlation coefficient matrix. The Pearson matrix is preferable, since it normalizes the element in a limited range of [-1,1], which unifies the element scale within the matrix and prevents overflowing. 
Second, as shown in Eq.~\ref{eq:lossBarlow}, Barlow Twins accumulated the error along the whole matrix $\phi$. However, the dimension of $\phi$ is large, \eg, $2048 \times 2048$. As training proceeds, the majority of elements converge, and the network arrives at the plateau, since the loss is accumulated by a vast of elements. The optimization of the remaining minority elements is usually ignored. 
The proposed DWDR also accumulated the error but with dynamic weights for different elements in Eq~\ref{eq:dwdr}.
We leverage the Pearson matrix, which is normalized in the range [-1, 1], to set the corresponding dynamic weights. The design also limits dynamic weights in [0,1], preventing the weight overflow. Therefore, DWDR can focus on the minority elements, even when majority channels are converged. Compared with Barlow Twins, DWDR encourages the network to make still correlated channels independent throughout the training period.


\textbf{Optimization.} 
We optimize our network by jointly employing the instance loss and DWDR:
\begin{equation}
L_{total} = \alpha L_{id} + (1 - \alpha)L_{DWDR}.
\label{eq:total}
\end{equation}
The instance loss $L_{id}$ forces different-platform images with the same geo-tag to be close on the high-level features and pushes mismatched images far apart.
At the same time, DWDR motivates the learned visual embeddings with independent channels. Thus the network is able to extract more discriminative features. $\alpha$ is a weight to control the importance of the loss function and the regularization term.

\section{Experiment} \label{experiment}
We introduce three cross-view geo-localization datasets and the evaluation protocol in Section~\ref{datasets}. The implementation detail is provided in Section~\ref{details}. We carry out a series of comparisons with state-of-the-art approaches in Section~\ref{comparsions}, followed by ablation studies in Section~\ref{ablations}. Finally, Section~\ref{visual} visualizes the cross-view geo-localization results.
\setlength{\tabcolsep}{1pt}
\subsection{Datasets and Evaluation Protocol}~\label{datasets}
We conduct experiments on three geo-localization datasets, \ie, University-1652~\citep{zheng_university-1652_nodate}, CVUSA~\citep{zhai_predicting_2017}, and CVACT~\citep{liu_lending_2019}. 
\par
\textbf{University-1652}~\citep{zheng_university-1652_nodate} is a multi-view multi-source dataset, including data from three different platforms, \ie, drones, satellites and dash cams. As the name implies, this dataset collects 1652 ordinary buildings of 72 universities around the world. 701 of all 1652 buildings are separated into the training set, and the other 951 builds constitute the testing set. Therefore, build images in the training and testing set are not overlapping. For each building, the dataset contains one satellite-view image, 54 drone-view images and 3.38 ground-view images on average. Since dash cams are hard to acquire enough street-view images for some buildings, the dataset also collects 21,099 common-view images from Google Image as an extra training set. The dataset supports two new aerial-view geo-localization tasks, \ie, drone-view target localization (Drone $\rightarrow$ Satellite) and drone navigation (Satellite $\rightarrow$ Drone). 
\par
\textbf{CVUSA}~\citep{zhai_predicting_2017} is a large-scale cross-view dataset, which consists of images from two viewpoints, \ie, the ground view and the satellite view. In the dataset, 35,532 ground-and-satellite image pairs are employed for training, and 8,884 image pairs are provided for testing. Noteworthily, ground-view images are the pattern of panoramas, and the orientation of all ground and satellite images is aligned.
\par
\textbf{CVACT}~\citep{liu_lending_2019} is a similar dataset to CVUSA. For the ground-to-satellite task, CVACT also contains 35,532 image pairs for training. Different from CVUSA, CVACT provides a validation set with 8,884 image pairs denoted as CVACT\_val. Meanwhile, compared with CVUSA, CVACT possesses a larger test set with 92,802 image pairs named CVACT\_test. When evaluated in CVACT\_val, a query ground-view panorama only match one satellite image in the gallery. However, for CVACT\_test, a panoramic query image may correspond to several satellite images within 5 meters from the ground-truth location. 
\par
We follow existing works~\citep{wang2021each,shi_optimal_nodate,shi_spatial-aware_nodate} and mainly employ CVACT\_val to evaluate our method when training on CVACT. The image number of training and evaluation in these three datasets is shown in Table~\ref{table:statistic}.

\setlength{\tabcolsep}{1pt}
\begin{table}[tb]
\small
\caption{
Statistics of three cross-view datasets, including the image number of training and testing sets. The left and right of the arrow ($\rightarrow$) refer to the query and gallery platforms, respectively.
}
\begin{center}
\resizebox{\linewidth}{!}{
\begin{tabular}{l|cc|cc|cc}
\hline
\multicolumn{1}{c|}{\multirow{2}{*}{Dataset}} & \multicolumn{2}{c|}{Training} &\multicolumn{4}{c}{Testing} \\
\cline{2-7}
& Drone & Satellite & \multicolumn{2}{c|}{Drone $\rightarrow$ Satellite} & \multicolumn{2}{c}{Satellite $\rightarrow$ Drone}\\
\shline
\makecell[c]{University\\-1652}~\citep{zheng_university-1652_nodate} & 37,854 & 701 & 37,855 & 951 & 701 & 51,355 \\
\hline
\cline{2-7}
& Ground & Satellite & \multicolumn{2}{c|}{Ground $\rightarrow$ Satellite} & \multicolumn{2}{c}{Satellite $\rightarrow$ Ground}\\

\shline
CVUSA~\citep{zhai_predicting_2017} & 35,532 & 35,532 & 8,884 & 8,884 & 8,884 & 8,884 \\
CVACT~\citep{liu_lending_2019} & 35,532 & 35,532 & 8,884 & 8,884 & 8,884 & 8,884 \\
\hline
\end{tabular}}
\end{center}
\label{table:statistic}
\end{table}

\par
\textbf{Evaluation protocol.} In our experiments, the performance of our method is evaluated by two commonly used metrics, \ie, Recall@K (\textbf{R@K}) and the average precision (\textbf{AP}). \textbf{R@K} refers to the proportion of true-matched candidates in the top-K of the ranking list. The value of \textbf{AP} is measured by the area under the Precision-Recall curve. Higher scores of these two metrics denote better performance of the network.
\par
\begin{table}[tb]
\small
\caption{
Comparison with the state-of-the-art results reported on University-1652. The compared method are categorized into three groups. The first group consists of baseline-related methods which employ average pooling to aggregate feature maps. The second group contains methods that apply contextual information. The third group includes Transformer-based methods. ``$M$'' indicates the margin of the triplet loss. $\dag$ denotes the input image of size $384 \times 384$. The input image size of two Transformer-based methods and other CNN-based methods are $224 \times 224$ and $256 \times 256$, respectively. The best results are in bold.} 
\begin{center}
\resizebox{\linewidth}{!}{
\begin{tabular}{l|cc|cc}
\hline
\multicolumn{1}{c|}{\multirow{3}{*}{Method}}  & \multicolumn{4}{c}{University-1652} \\
\cline{2-5}
& \multicolumn{2}{c|}{Drone $\rightarrow$ Satellite} & \multicolumn{2}{c}{Satellite $\rightarrow$ Drone}\\
& R@1 & AP & R@1 & AP \\
\shline
Instance Loss (\textbf{Baseline}) ~\citep{zheng_university-1652_nodate} & 57.09 & 61.88 & 73.89 & 58.73 \\
Contrastive Loss~\citep{lin_learning_2015} & 52.39 & 57.44 & 63.91 & 52.24 \\
Triplet Loss ($M=0.3$)~\citep{chechik2009large} & 52.16 & 57.47 & 65.05 & 52.37 \\
Triplet Loss ($M=0.5$)~\citep{chechik2009large} & 51.23 & 56.40 & 62.77 & 51.29 \\
Soft Margin Triplet Loss~\citep{hu_cvm-net_2018} & 53.67 & 58.69 & 67.90 & 54.76 \\
LCM$^\dag$~\citep{ding2021practical} & 66.65 & 70.82 & 79.89 & 65.38 \\
RK-Net~\citep{lin2022joint} & 66.13 & 70.23 & 80.17 & 65.76 \\
\textbf{Baseline~\citep{zheng_university-1652_nodate} + Ours} & \textbf{69.77} & \textbf{73.73} & \textbf{81.46} & \textbf{70.45} \\
\hline
LPN~\citep{wang2021each} & 75.93 & 79.14 & 86.45 & 74.49 \\
LPN + USAM~\citep{lin2022joint} & 77.60 & 80.55 & 86.59 & 75.96 \\
PCL~\citep{tian2021pcl} & 79.47 & 83.63 & 87.69 & 78.51 \\
\textbf{LPN~\citep{wang2021each} + Ours} & \textbf{81.51} & \textbf{84.11} & \textbf{88.30} & \textbf{79.38} \\
\hline
Swin-B~\citep{liu2021swin} & 84.15 & 86.62 & 90.30 & 83.55 \\
FSRA~\citep{dai2021transformer} & 84.51 & 86.71 & 88.45 & 83.47 \\
\textbf{Swin-B~\citep{liu2021swin} + Ours} & \textbf{86.41} & \textbf{88.41} & \textbf{91.30} & \textbf{86.02} \\
\hline
\end{tabular}}
\end{center}
\label{table:university1652}
\end{table}
\subsection{Implementation Details}~\label{details}
Our method is performed based on a classic two-branch baseline~\citep{zheng_university-1652_nodate}. The baseline adopts a modified ResNet-50~\citep{he2016deep} pre-trained on ImageNet~\citep{5206848} to extract visual features. Specifically, we remove the final classification layer of ResNet-50~\citep{he2016deep}. Besides, the stride of the second convolution layer and the down-sample layer in the first bottleneck of the ResNet-50~\citep{he2016deep} stage4 is set from 2 to 1. The input image is resized to $256 \times 256$, and the image augmentation consists of random cropping, random horizontal flipping, and random rotation. We employ stochastic gradient descent (SGD) with
momentum 0.9 and weight decay 0.0005 to update model parameters. The image number of each platform in a mini-batch is 16. 
The initial learning rate is 0.001 for the modified ResNet-50~\citep{he2016deep} backbone and 0.01 for the classifier module. The dropout rate in the classifier module is 0.75. We train the two-branch baseline for 120 epochs, and the learning rate is decayed by 0.1 after 80 epochs. There are two trade-off parameters $\lambda$ and $\alpha$ in the loss function. We run a simple search and observe the better results for $\lambda=1.3 \times 10^{-3}$ and $\alpha=0.9$. Note that when using Swin-B~\citep{liu2021swin} and VGG16~\citep{vgg} as backbones, $\lambda=2.0 \times 10^{-3}$ and $\lambda=3.9 \times 10^{-3}$ are best choices, separately. During testing, we deploy the Euclidean distance to compute the similarities between the query and candidates. Our model is implemented on Pytorch~\citep{paszke2019pytorch}, and all experiments are conducted on a single NVIDIA RTX 2080Ti GPU.
\par

\begin{table*}[tb]
\setlength{\tabcolsep}{7pt}
\centering
\small
\caption{
Comparison with prior art on CVUSA and CVACT. The compared methods are divided into 2 columns. Column1: methods without the polar transform. Column2: methods utilizing the polar transform. ``Polar Transform'' is the boundary of two group Columns. $\ddag$: The method is implemented using images processed by the polar transform. $\star$: The method harnesses extra orientation information as input. The best results are in bold.
}
\resizebox{\linewidth}{!}{
\begin{tabular}{l|c|c|cccc|cccc}
\hline
\multicolumn{1}{c|}{\multirow{2}{*}{Method}} & \multirow{2}{*}{Publication} & \multirow{2}{*}{Backbone} & \multicolumn{4}{c|}{CVUSA} & \multicolumn{4}{c}{CVACT\_val}\\ 
\cline{4-11}
                                        & & & R@1 & R@5 & R@10 & R@Top1\% & R@1 & R@5 & R@10 & R@Top1\% \\
\shline
MCVPlaces~\citep{workman_wide-area_2015} & ICCV'15 & AlexNet & - & - & - & 34.40 & - & - & - & - \\
Zhai~\citep{zhai_predicting_2017} & CVPR'17 & VGG16 & - & - & - & 43.20 & - & - & - & -\\
Vo~\citep{vo_localizing_2017}    & ECCV'16 & AlexNet & - & - & - & 63.70 & - & - & - & -\\
CVM-Net~\citep{hu_cvm-net_2018} & CVPR'18 & VGG16 & 18.80 & 44.42 & 57.47 & 91.54 & 20.15 & 45.00 & 56.87 & 87.57\\
Orientation$^\star$~\citep{liu_lending_2019}   & CVPR'19 & VGG16 & 27.15 & 54.66 & 67.54 & 93.91 & 46.96 & 68.28 & 75.48 & 92.04\\
Zheng~(\textbf{Baseline})~\citep{zheng_university-1652_nodate} & MM'20 & VGG16 & 43.91 & 66.38 & 74.58 & 91.78 & 31.20 & 53.64 & 63.00 & 85.27\\
Regmi~\citep{Regmi_2019_ICCV} & ICCV'19 & X-Fork & 48.75 & - & 81.27 & 95.98 & - & - & - & -\\
RKNet~\citep{lin2022joint} & TIP'22 & USAM & 52.50 & - & - & 96.52 & 40.53 & - & - & 89.12\\
Siam-FCANet~\citep{Siam-FCANet} & ICCV'19 & ResNet-34 & - & - & - & 98.30 & - & - & - & -\\
CVFT~\citep{shi_optimal_nodate} & AAAI'20 & VGG16 & 61.43 & 84.69 & 90.94 & 99.02 & 61.05 & 81.33 & 86.52 & 95.93\\
LPN~\citep{wang2021each} & TCSVT'21 & ResNet-50 & 85.79 & 95.38 & 96.98 & 99.41 & 79.99 & 90.63 & 92.56 & 97.03\\
LPN + USAM~\citep{lin2022joint} & TIP'22 & ResNet-50 & 91.22 & - & - & 99.67 & 82.02 & - & -& \textbf{98.18} \\
\arrayrulecolor{gray} \hline
\multicolumn{11}{c}{Polar Transform} \\
\arrayrulecolor{gray}  \hline
SAFA~\citep{shi_spatial-aware_nodate} & NIPS'19 & VGG16 & 89.84 & 96.93 & 98.14 & 99.64 & 81.03 & \textbf{92.80} & \textbf{94.84} & 98.17\\
DSM~\citep{shi2020looking} & CVPR'20 & VGG16 & 91.96 & 97.50 & 98.54 & 99.67 & 82.49 & 92.44 & 93.99 & 97.32 \\
LPN$^\ddag$~\citep{wang2021each} & TCSVT'21 & ResNet-50 & 93.78 & 98.50 & 99.03 & 99.72 & 82.87 & 92.26 & 94.09 & 97.77 \\
\hline
\textbf{Baseline + Ours} & - & VGG16 & 75.62 & 90.45 & 93.60 & 98.60 & 66.76 & 83.34 & 87.11 & 95.10\\
\textbf{LPN$^\ddag$~\citep{wang2021each} + Ours} & - & ResNet-50 & \textbf{94.33} & \textbf{98.54} & \textbf{99.09} & \textbf{99.80} & \textbf{83.73} & 92.78 & 94.53 & 97.78 \\
\hline
\end{tabular}}
\label{table:CVUSA}
\end{table*}

\subsection{Comparison with State-of-the-art Methods}~\label{comparsions}
\textbf{Results on University-1652.} As shown in Table~\ref{table:university1652}, we compare our method with lots of competitive methods on University-1652. The compared methods are divided into three groups, \ie, baseline-related methods, methods harnessing contextual information and Transformer-based methods. In the first group, the first line reports results of our two-branch baseline, \ie, ``Instance Loss~\citep{zheng_university-1652_nodate}''. We can observe that ``Baseline + Ours'' substantially improves the baseline performance. In the drone-view target localization task (Drone $\rightarrow$ Satellite), the accuracy of R@1 increases from $57.09\%$ to $69.77\%$ ($+12.68\%$), and the value of AP raises from $61.88\%$ to $73.73\%$ ($+11.85\%$). In the drone navigation task (Satellite $\rightarrow$ Drone), the accuracy of R@1 goes up from $73.89\%$ to $81.46\%$ ($+7.57\%$), and the value of AP increases from $58.73\%$ to $70.45\%$ ($+11.72\%$). Meanwhile, the performance of our method also has surpassed other baseline-related methods. In the second group, LPN~\citep{wang2021each} explicitly takes advantage of contextual information during training. Some methods, \eg, ``LPN + USAM~\citep{lin2022joint}'' and PCL~\citep{tian2021pcl}, combined with LPN have yielded better results, and we can also implement our method based on LPN. Specifically, we re-implement LPN by utilizing the symmetric sampling strategy to replace the original random sampling and incorporating the dynamic weighted decorrelation regularization during training. Compared with results of LPN,
``LPN + Ours'' achieves $81.51\%$ R@1 accuracy ($+5.58\%$) and $84.11\%$ AP ($+4.97\%$) on Drone $\rightarrow$ Satellite and $88.30\%$ R@1 accuracy ($+1.85\%$) and $79.38\%$ AP ($+4.89\%$) on Satellite $\rightarrow$ Drone. 
The feature expression ability of Transformer is stronger than that of CNN, and both Transformer-based methods~\citep{liu2021swin, dai2021transformer} obtain a better performance than CNN-based methods. We further combine our method with ``Swin-B~\citep{liu2021swin}''. ``Swin-B'' indicates the two-branch baseline applying Swin-B as the backbone. ``Swin-B + Ours'' on University-1652 achieves the state-of-the-art results, \ie, $86.41\%$ in R@1 accuracy and $88.41\%$ in AP for Drone $\rightarrow$ Satellite and $91.30\%$ in R@1 accuracy and $86.02\%$ in AP for Satellite $\rightarrow$ Drone.
\par
\textbf{Results on CVUSA and CVACT.}
Comparisons with other competitive approaches on CVUSA and CVACT are summarized in Table~\ref{table:CVUSA}. CVUSA and CVACT have a similar data pattern, \ie, the satellite-platform images of aerial viewpoint and the ground panoramas. The polar transform considers the geometric correspondence of two-platform images and transforms the aerial-view image to approximately align a ground panorama at the pixel level. The aligned images help to improve the performance of models. Depending on whether or not the polar transform is harnessed, the compared method can be divided into two columns.
The first column reports methods without using polar transform, and methods in the second column employ the polar transform during training and testing. Our method does not employ the polar transform. Experiments on CVUSA and CVACT show phenomena similar to that on University-1652. Our method first outperforms a dual-stream baseline (\ie, the method of Zheng~\citep{zheng_university-1652_nodate}) by a large margin, \ie, $31.71\%$ R@1 improvement on CVUSA and $35.56\%$ R@1 raising on CVACT. At the same time, our method exceeds most of existing methods in the first column. In particular, our method obtains $75.62\%$ R@1, $90.45\%$ R@5, $93.60\%$ R@10, and $98.60\%$ R@Top1\% on CVUSA, and $66.76\%$ R@1, $83.34\%$ R@5, $87.11\%$ R@10, and $95.10\%$ R@Top1\% on CVACT. 
In experiments of University-1652, we observe that our method can combine with LPN~\citep{wang2021each} and achieve better results. The same experiments are also carried out on CVUSA and CVACT. There are two versions of LPN (\ie, LPN and LPN$^\ddag$) in Table~\ref{table:CVUSA}. LPN$^\ddag$ applies the polar transform and has achieved higher performance. We notice that our approach still yields competitive results when complemented with the LPN$^\ddag$. For instance, ``LPN$^\ddag$ + Ours'' boosts the R@1 accuracy from $93.78\%$ to $94.33\%$ on CVUSA and $82.87\%$ to $83.73\%$ on CVACT. 
\par
The above experimental results on three cross-view geo-localization datasets suggest two points. One is that our method can be flexibly applied in different cross-view settings. The other is that our method is able to encourage existing approaches to mine more diverse patterns, yielding discriminative features.

\subsection{Ablation Studies}~\label{ablations}
To further analyze our method, we design several ablation studies. The ablation studies are mainly based on the drone-view target localization (Drone $\rightarrow$ Satellite) and drone navigation (Satellite $\rightarrow$ Drone) of University-1652~\citep{zheng_university-1652_nodate}. 
\par
\textbf{Analysis of parameters $\gamma_{1}$ and $\gamma_{2}$.} The main contribution of our paper is the proposed dynamic weighted decorrelation regularization (DWDR). In DWDR, $\gamma_{1}$ and $\gamma_{2}$ are two key parameters that flexibly adjust the rate at which well-regressed elements of the Pearson cross-correlation coefficient matrix are down-weighted. When $\gamma_{1} = 0$ and $\gamma_{2} = 0$, DWDR does not apply two dynamic weights and can be viewed as Barlow Twins~\citep{zbontar2021barlow}. We empirically tune $\gamma_{1}$ and $\gamma_{2}$, and the related results are detailed in Table~\ref{weight Ablation}. We first observe that applying one dynamic weight, \ie, only $\gamma_{1} = 1$ or only $\gamma_{2} = 1$, achieves similar results to Barlow Twins. The limited performance improvement reflects that ignoring poorly-regressed diagonal and off-diagonal elements both induce the optimization plateau. When both $\gamma_{1}$ and $\gamma_{2}$ are set to 1, \ie, using two dynamic weights, we obtain the best results. Specifically, compared with deploying Barlow Twins as regularization (``Baseline + \textit{BT}''), our method (``Baseline + Ours'') boosts R@1 from $67.91\%$ to $69.77\%$ ($+1.86\%$) and AP from $71.99\%$ to $73.73\%$ ($+1.74\%$) on Drone $\rightarrow$ Satellite, and goes up R@1 from $80.17\%$ to $81.46\%$ ($+1.29\%$) and AP from $68.03\%$ to $70.45\%$ ($+2.42\%$) on Satellite $\rightarrow$ Drone. When $\gamma_{1} = 2$ and $\gamma_{2} = 2$, the performance gains slightly degrades. A reasonable speculation is that large \textit{focusing} parameters $\gamma_{1}$ and $\gamma_{2}$ cause the importance of poorly-regressed elements in the optimization process to be excessively reduced as well. To further verify the robustness of selected parameters, we conduct the same experiments in ``LPN~\citep{wang2021each} + Ours'' and ``Swin-B~\citep{liu2021swin} + Ours'' and find the same conclusion. That is, when both $\gamma_{1}$ and $\gamma_{2}$ are set to 1, models achieve competitive results. Therefore, we choose $\gamma_{1} = 1$ and $\gamma_{2} = 1$ as default \textit{focusing} parameters of DWDR. All three groups of experiments also support that DWDR is more effective than Barlow Twins for motivating networks to learn low-redundancy visual embeddings.
\par
\begin{table}[tb]
\setlength{\tabcolsep}{3pt}
\centering
\small
\caption{Ablation study with different $\gamma_{1}$ and $\gamma_{2}$ in the dynamic weighted decorrelation regularization. \textit{BT} refers to Barlow Twins~\citep{zbontar2021barlow}.}
\label{weight Ablation}
\resizebox{\linewidth}{!}{
\begin{tabular}{l|cc|cc|cc}
\shline
\multicolumn{1}{c|}{\multirow{2}{*}{Method}} & \multirow{2}{*}{$\gamma_{1}$} & \multirow{2}{*}{$\gamma_{2}$} & \multicolumn{2}{c|}{Drone $\rightarrow$ Satellite} & \multicolumn{2}{c}{Satellite $\rightarrow$ Drone} \\ 
\cline{4-7} 
 & & & R@1 & AP & R@1 & AP \\ 
\hline
Baseline~\citep{zheng_university-1652_nodate}+\textit{BT} & 0 & 0 & 67.91  &  71.99  &  80.17  &  68.03 \\
\hline
\multirow{4}{*}{Baseline~\citep{zheng_university-1652_nodate}+Ours} 
  & 0 & 1 &  66.83  &  71.01  &  77.89  & 68.01  \\
 & 1 & 0 &  67.57  &  71.62  &  78.03  & 67.93  \\
 & 1 & 1  & \textbf{69.77} &  \textbf{73.73} & \textbf{81.46} & \textbf{70.45} \\ 
 & 2 & 2  & 69.40 &  73.33 & 80.88 & 70.05 \\
\hline
LPN~\citep{wang2021each}+\textit{BT} & 0 & 0 & 80.93 & 83.60 & 86.02 & 78.33   \\
\hline
\multirow{4}{*}{LPN~\citep{wang2021each}+Ours}
 & 0 & 1 &  80.84  &  83.50  &  87.30  & 79.26  \\
 & 1 & 0 &  80.83  &  83.49  &  88.30  & \textbf{79.93}  \\
 & 1 & 1  & \textbf{81.51} &  \textbf{84.11} & 88.30 & 79.38 \\ 
 & 2 & 2  & 80.49 &  83.17 & \textbf{88.45} & 79.91 \\
\hline
Swin-B~\citep{liu2021swin}+\textit{BT} & 0 & 0 & 86.03 & 88.05 & 91.01 & 85.07   \\
\hline
\multirow{4}{*}{Swin-B~\citep{liu2021swin}+Ours}
 & 0 & 1 &  85.94  &  88.00  &  91.01  & 85.33  \\
 & 1 & 0 &  86.07  &  88.09  &  90.30  & 85.68  \\
 & 1 & 1  & \textbf{86.41} &  \textbf{88.41} & \textbf{91.30} & \textbf{86.02} \\ 
 & 2 & 2  & 85.54 &  87.73 & 90.58 & 85.65 \\
\hline
\end{tabular}}
\end{table}

\textbf{Effect of our sampling strategy and DWDR.}
Our symmetric sampling strategy is a combination of the drone-view based sampling and the satellite-view based sampling. To discuss the effectiveness of our sampling strategy, we conduct three groups of experiments under the condition of only changing the sampling strategy. Meanwhile, in each group of experiments, we study the effectiveness of DWDR. The experimental results are shown in Table~\ref{structures Ablation}. We observe first that utilizing DWDR alone does not give comparable results to the baseline (``Instance Loss~\citep{zheng_university-1652_nodate}'') shown in Table~\ref{table:university1652}. However, when applied in conjunction with Instance Loss, DWDR significantly improves the performance of the network, regardless of the sampling strategy. Experiments within each group verify from the side that DWDR concentrates more on the redundant channels of the embedding itself rather than the distance between cross-view embeddings. Second, when only Instance Loss is harnessed, the drone-view based and the satellite-view based sampling acquire similar results to the baseline (``Instance Loss'') applying random sampling. In contrast, the symmetric sampling strategy obtains the best geo-localization accuracy. Furthermore, the symmetric sampling strategy is also the most competitive in the other two experimental settings, \ie, DWDR alone and Instance Loss plus DWDR. The significant performance increment demonstrates that the symmetric sampling strategy as a by-product is effective.

\begin{table}[!t]
\renewcommand{\arraystretch}{1.3}
\caption{Effect of the symmetric sampling strategy and the dynamic weighted decorrelation regularization (DWDR).}
\label{structures Ablation}
\centering
\resizebox{1.0\linewidth}{!}{
\begin{tabular}{l|cc|cc|cc}
\shline
\multicolumn{1}{c|}{\multirow{2}{*}{Method}} & \multirow{2}{*}{\makecell[c]{Instance\\Loss}} & \multirow{2}{*}{DWDR} & \multicolumn{2}{c|}{Drone $\rightarrow$ Satellite} & \multicolumn{2}{c}{Satellite $\rightarrow$ Drone} \\ 
\cline{4-7} 
  & & & R@1 & AP & R@1 & AP \\ 
\hline
\multirow{3}{*}{\makecell[c]{Drone-view\\based sampling}} & \textbf{ \checkmark } &  & 60.86  &  65.56  &  73.89   &  59.05   \\
 &  & \textbf{ \checkmark }  &  22.61  &  27.84  &  35.38  & 20.98  \\
 & \textbf{ \checkmark } & \textbf{ \checkmark } & \textbf{66.08} &  \textbf{70.22} & \textbf{77.60} & \textbf{65.48} \\ 
\hline
\multirow{3}{*}{\makecell[c]{Satellite-view\\based sampling}} & \textbf{ \checkmark } &  & 58.54 & 63.10 & 73.61 & 58.49   \\
 &  & \textbf{ \checkmark } &  24.25  &  29.37  &  39.09  & 22.83  \\
 & \textbf{ \checkmark } & \textbf{ \checkmark }  & \textbf{65.06} &  \textbf{69.16} & \textbf{76.46} & \textbf{65.27} \\ 
\hline
\multirow{3}{*}{\makecell[c]{The symmetric\\sampling strategy\\(\textbf{Ours})}} & \textbf{ \checkmark } &  & 64.74 & 68.96 & 77.75 & 64.32   \\
 &  & \textbf{ \checkmark } &  34.38  &  40.02  &  50.07  & 34.28  \\
 & \textbf{ \checkmark } & \textbf{ \checkmark }  & \textbf{69.77} &  \textbf{73.73} & \textbf{81.46} & \textbf{70.45} \\ 
\hline
\end{tabular}}
\end{table}

\begin{figure}[tb]
  \centering
  \includegraphics[width=1\linewidth]{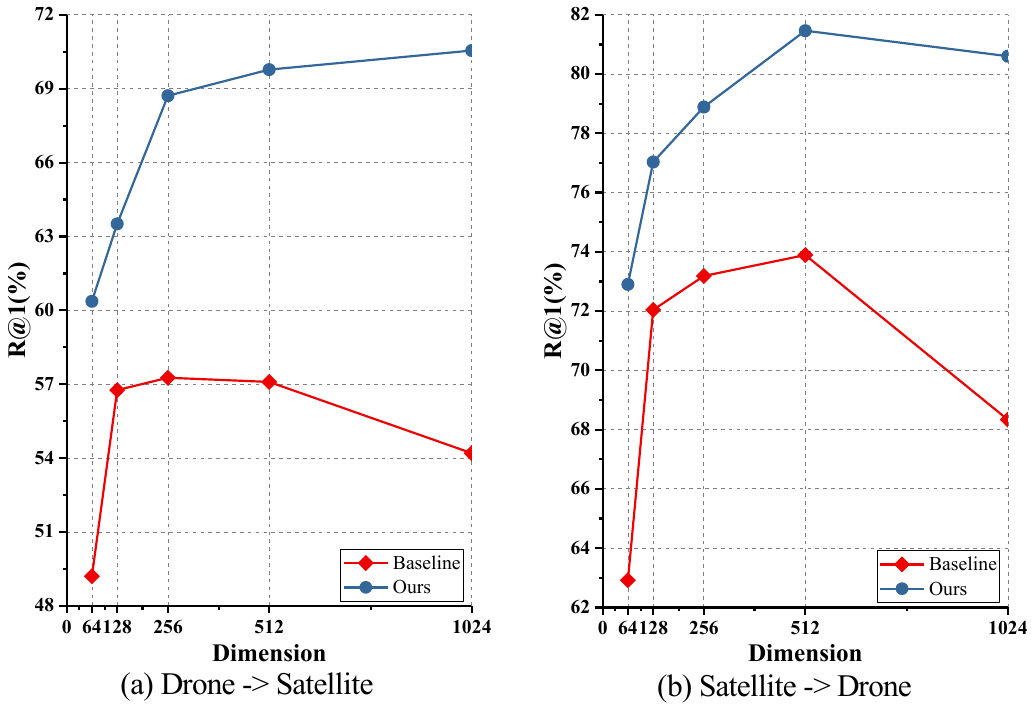}
  \caption{Impact of the dimension of features. The R@1 accuracy between the baseline and our method is compared. The \textcolor[RGB]{255,0,0}{red line} refers to the baseline~\citep{zheng_university-1652_nodate}, and our method is shown using the \textcolor[RGB]{0,0,255}{blue line}. (a) The drone-view target localization task (Drone $\rightarrow$ Satellite). (b) The drone navigation task (Satellite $\rightarrow$ Drone). When the feature dimension changes from 128 to 64, the performance of our method drops less than the baseline. 
  }
  \label{fig:dim}
\end{figure}

\textbf{Effect of the dimension of visual embeddings.}
We deploy the final visual embeddings with different dimensions in geo-localization to investigate the effect of embedding dimensions on retrieval accuracy. The experimental results of the baseline and ``Baseline + Ours'' are shown in Table~\ref{dim Ablation}. We observe that with the increment of the dimension, both the baseline~\citep{zheng_university-1652_nodate} and ``Baseline + Ours'' have a persistent improvement since the visual embedding possesses more information capacity. However, the performance of the two methods encounters the bottleneck when the feature dimension is $512$. As the dimension of the feature increases to $1024$, the performance of the baseline decreases significantly, and the performance of ``Baseline + Ours'' tends to stabilize. The experimental results reflect two aspects from the side. One is that features with too high dimensions are prone to redundant channels, which compromise the geo-localization accuracy of models. The other is that our method can encourage networks to learn low-redundancy embeddings and improve the robustness of the model. In addition, as shown in Figure~\ref{fig:dim}, we notice that when the dimension raises from $64$ to $128$, the baseline achieves a higher growth rate than ``Baseline + Ours''. The short-dimensional features with small information capacity limit the performance of models. We speculate that our method allows the model to include more primary discriminative patterns in the limited feature dimension to mitigate the negative effects of insufficient information capacity. Therefore, when the feature dimension increases, our method produces fewer performance fluctuations.
\begin{table}[!t]
\setlength{\tabcolsep}{3pt}
\centering
\small
\caption{Ablation study of cross-view geo-localization applying visual features with different dimensions. ``Dim'' denotes the dimension of features.}
\label{dim Ablation}
\resizebox{\linewidth}{!}{
\begin{tabular}{l|c|cc|cc}
\shline
\multicolumn{1}{c|}{\multirow{2}{*}{Method}} & \multirow{2}{*}{Dim} & \multicolumn{2}{c|}{Drone $\rightarrow$ Satellite} & \multicolumn{2}{c}{Satellite $\rightarrow$ Drone} \\ 
\cline{3-6} 
 & & R@1 & AP & R@1 & AP \\ 
\hline
\multirow{5}{*}{\makecell[c]{Baseline~\citep{zheng_university-1652_nodate}\\(Instance Loss)}} & 64 & 49.20  &  54.36  &  62.91   &  49.68   \\
  & 128 &  56.76  &  61.74  &  72.04  & 58.14  \\
 & 256 &  \textbf{57.26}  &  \textbf{62.17}  &  73.18  & 58.70  \\
 & 512  & 57.09 &  61.88 & \textbf{73.89} & \textbf{58.73} \\ 
 & 1024 & 54.20 & 59.20 & 68.33 & 55.37 \\
\hline
\multirow{5}{*}{Baseline~\citep{zheng_university-1652_nodate}+Ours} & 64 & 60.37 & 65.03 & 72.90 & 60.31   \\
 & 128 &  63.51  &  68.05  &  77.03  & 64.57  \\
 & 256 &  68.71  &  72.72  &  78.89  & 68.44  \\
 & 512 & 69.77 &  73.73 & \textbf{81.46} & 70.45 \\ 
 & 1024 & \textbf{70.55} &  \textbf{74.56} & 80.60 & \textbf{70.51} \\
 \hline
\end{tabular}}
\end{table}

\textbf{Effect of DWDR under different loss functions.}
Our baseline applies the instance loss~\citep{zheng2020dual,zheng2017unlabeled} to optimize the network while other loss functions are available. The triplet loss and the soft margin triplet loss are broadly utilized in previous works~\citep{hu_cvm-net_2018,liu_lending_2019,shi_optimal_nodate,vo_localizing_2017}. We also evaluate our DWDR by deploying baselines adopting these two loss functions. The margin value of the triplet loss is 0.3, and experimental results are shown in Table~\ref{table:triplet}. We notice that both baselines combined with DWDR gain improved retrieval accuracy on the ``Drone $\rightarrow$ Satellite'' task and the ``Satellite $\rightarrow$ Drone'' task of University-1652.

\textbf{Effect of the intra-view DWDR.}
Our method applies the cross-view DWDR, in which the Pearson cross-correlation coefficient matrix is computed employing cross-view images. The intra-view DWDR means that the Pearson cross-correlation coefficient matrix of DWDR is calculated employing two distorted images from the same platform generated by different data augmentations. In experiments, the method only utilizing the symmetric sampling strategy is treated as the baseline, and the comparison results are shown in Table~\ref{table:intra dwdr}. We observe that the baseline combined with the intra-view DWDR gains a slight increment. Although the intra-view DWDR also encourages the network to learn independent embedding channels, our cross-view DWDR significantly outperforms applying the intra-view DWDR. It is because the cross-view DWDR is aligned with the cross-view retrieval test setting, which considers embeddings from different platforms for the geo-localization task. It also explains the limited performance increase of applying both intra-view and cross-view DWDR, which relies on the cross-view DWDR. 

\begin{table}[tb]
\small
\caption{Ablation study of DWDR under different loss functions. ``$M$'' denotes the margin of the triplet loss.} 
\begin{center}
\resizebox{\linewidth}{!}{
\begin{tabular}{l|cc|cc}
\hline
\multicolumn{1}{c|}{\multirow{3}{*}{Method}}  & \multicolumn{4}{c}{University-1652} \\
\cline{2-5}
& \multicolumn{2}{c|}{Drone $\rightarrow$ Satellite} & \multicolumn{2}{c}{Satellite $\rightarrow$ Drone}\\
& R@1 & AP & R@1 & AP \\
\shline
Triplet Loss ($M=0.3$)~\citep{chechik2009large} & 52.16 & 57.47 & 63.91 & 52.24 \\
Soft Margin Triplet Loss~\citep{hu_cvm-net_2018} & 53.67 & 58.69 & 67.90 & 54.76 \\
\shline
Triplet Loss ($M=0.3$) + DWDR & 54.14 & 59.28 & 67.90 & 54.76 \\
Soft Margin Triplet Loss + DWDR & 57.75 & 62.58 & 69.33 & 57.46 \\
\hline
\end{tabular}}
\end{center}
\label{table:triplet}
\end{table}

\setlength{\tabcolsep}{1pt}
\begin{table}[tb]
\small
\caption{Ablation study of the symmetric sampling strategy combined with different DWDR.
} 
\begin{center}
\resizebox{\linewidth}{!}{
\begin{tabular}{l|cc|cc}
\hline
\multicolumn{1}{c|}{\multirow{3}{*}{Method}}  & \multicolumn{4}{c}{University-1652} \\
\cline{2-5}
& \multicolumn{2}{c|}{Drone $\rightarrow$ Satellite} & \multicolumn{2}{c}{Satellite $\rightarrow$ Drone}\\
& R@1 & AP & R@1 & AP \\
\shline
Symmetric sampling~\textbf{(Baseline)} & 64.74 & 68.96 & 77.75 & 64.32 \\
\hline
+ Intra-view DWDR & 65.31 & 69.57 & 79.17 & 65.74 \\
+ Cross-view DWDR & 69.77 & 73.73 & 81.46 & 70.45 \\
+ Intra \& Cross-view DWDR & 69.81 & 73.68 & 82.45 & 70.86 \\
\hline
\end{tabular}}
\end{center}
\label{table:intra dwdr}
\end{table}


\subsection{Qualitative Results}~\label{visual}
\begin{figure}[tb]
  \centering
  \includegraphics[width=1\linewidth]{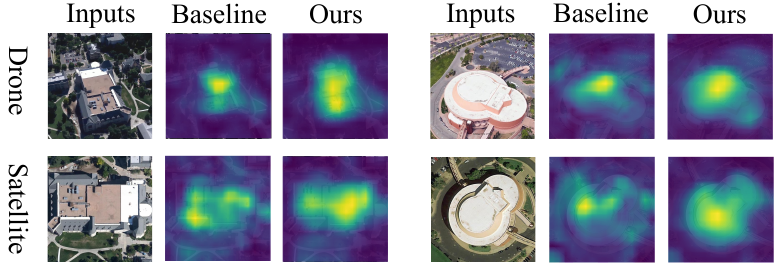}
  \caption{Visualization of heatmaps. Heatmaps are produced by the baseline~\citep{zheng_university-1652_nodate} and ours on different platforms of University-1652, \ie, the drone platform and the satellite platform. 
  }
  \label{fig:heatmap}
\end{figure}

\begin{figure}[tb]
  \centering
  \includegraphics[width=0.95\linewidth]{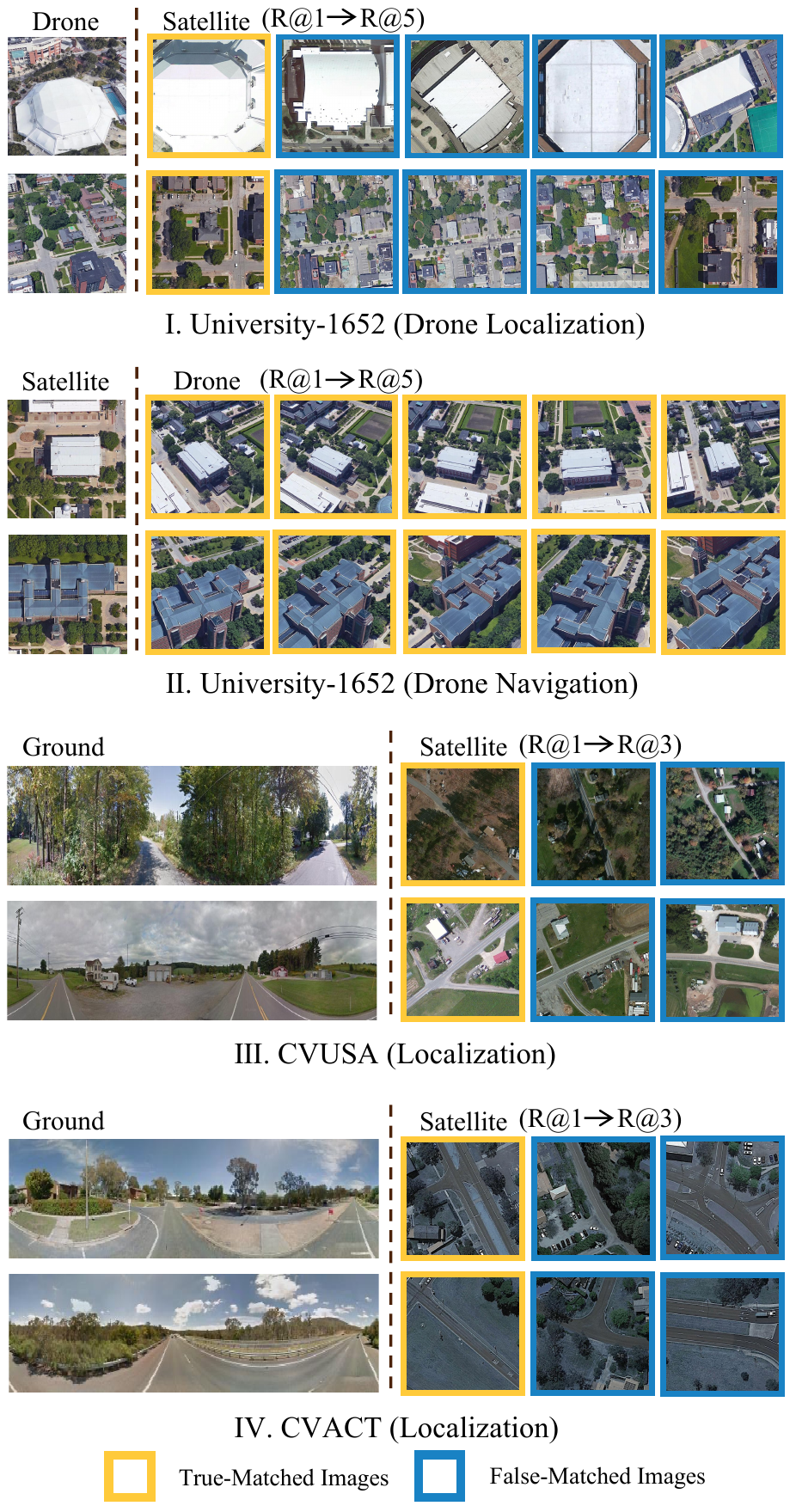}
  \caption{Qualitative image retrieval results in different datasets. (I) and (II) show Top-5 retrieval results on University-1652. Different query images indicate the different tasks. (I) is the drone-view target localization task, and (II) is the drone navigation task. (III) and (IV) exhibit Top-3 retrieval results of geographic localization on CVUSA and CVACT, respectively. The true matches are in yellow boxes, and the false matches are highlighted by blue boxes. 
  }
  \label{fig:rank}
\end{figure}

We visualize some heatmaps generated by the baseline and our method as an extra qualitative evaluation. Figure~\ref{fig:heatmap} shows the acquired heatmaps in the drone and satellite platforms of University-1652. Images in University-1652 possess an obvious geographic target. Compared with the baseline~\citep{zheng_university-1652_nodate}, our method activates a wider range of geographic target regions. In addition, we show some retrieval results on different datasets (see Figure~\ref{fig:rank}). University-1652 supports two tasks. In the drone-view target localization task, the drone-platform image is the query, and in the drone navigation task, the satellite-platform image is the query. The retrieval results of two tasks are shown in Figure~\ref{fig:rank} (I) and (II). Figure~\ref{fig:rank} (III) and (IV) show the retrieval results of the ground-to-satellite localization task on CVUSA and CVACT. Given a randomly selected test query, we notice that the proposed method has successfully retrieved the most relevant results from the candidate gallery. 

\section{Conclusion}~\label{conclusion}
In this paper, we propose a dynamic weighted decorrelation regularization (DWDR) to achieve the cross-view geo-localization. DWDR reduces the redundancy of visual embeddings by motivating the network to learn independent embedding channels. 
Specifically, DWDR sets dynamic weights to focus on the poorly-regressed elements when constraining the objective matrix to be as close as possible to the identity matrix. 
As a by-product of DWDR, the cross-view symmetric sampling strategy is introduced to balance the example number from different platforms in a training batch. The extensive experiments on three datasets, \ie, University-1652, CVUSA and CVACT, demonstrate that our method can learn discriminative embeddings, which significantly improve the retrieval accuracy. Moreover, our method also acquires competitive results with the extremely short feature.

\section*{Data Availability Statement}
Three datasets supporting the findings of this study are available with the permission of the dataset authors. The links to request these datasets are as follows. \\(1) University-1652 : \href{https://github.com/layumi/University1652-Baseline}{https://github.com/layumi/University16\\52-Baseline};\\ (2) CVUSA : \href{https://github.com/viibridges/crossnet}{https://github.com/viibridges/crossnet}; \\(3) CVACT : \href{https://github.com/Liumouliu/OriCNN}{https://github.com/Liumouliu/OriCNN}.


{\footnotesize
\bibliographystyle{apalike}
\bibliography{ieee}
}

\end{document}